\pdfoutput=1

\documentclass[11pt]{article}

\usepackage[final]{acl}

\usepackage{times}
\usepackage{latexsym}
\usepackage[T1]{fontenc}
\usepackage{bm}
\usepackage[utf8]{inputenc}

\usepackage{microtype}

\usepackage{inconsolata}

\usepackage{graphicx}

\usepackage{url}            
\usepackage{booktabs}       
\usepackage{amsfonts}       
\usepackage{nicefrac}       
\usepackage{xcolor}         
\usepackage{amsmath}
\usepackage{amsthm}
\usepackage{amssymb}
\usepackage{bbm}
\usepackage{pifont}
\usepackage{multirow}
\usepackage{subcaption}
\usepackage{cleveref}
\usepackage{array}
\usepackage{framed}
\usepackage{listings}
\usepackage{todonotes}

\newcommand{\mytilde}{\raise.17ex\hbox{$\scriptstyle\mathtt{\sim}$}}

\definecolor{OliveGreen}{rgb}{0,0.6,0}

\newcommand{\mistralinstruct}{\textit{{Mistral-7B-Instruct-v0.3}}}
\newcommand{\qwenthinking}{\textit{{Qwen3-32B}}}
\newcommand{\qwenbase}{\textit{{Qwen3-14B-Base}}}
\newcommand{\llamabig}{\textit{{Llama-3.3-70B-Instruct}}}

\title{LLMs Underperform Graph-Based Parsers \\ on Supervised Relation Extraction for Complex Graphs}

\author{%
  Paolo Gajo
    \\
  University of Bologna\\
  \texttt{paolo.gajo2@unibo.it} \\
  \And
  Domenic Rosati \\
  Dalhousie University \\
  \texttt{Domenic.Rosati@Dal.Ca} \\
  \AND
  Hassan Sajjad \\
  Dalhousie University \\
  \texttt{HSajjad@dal.ca} \\
  \And
  Alberto Barrón-Cedeño \\
  University of Bologna \\
  \texttt{a.barron@unibo.it} \\
}

\begin{document}
\maketitle
\begin{abstract}
Relation extraction represents a fundamental component in the process of creating knowledge graphs, among other applications. Large language models (LLMs) have been adopted as a promising tool for relation extraction, both in supervised and in-context learning settings.
However, in this work we show that their performance still lags behind much smaller architectures when the linguistic graph underlying a text has great complexity.
To demonstrate this, we evaluate four LLMs against a graph-based parser on six relation extraction datasets with sentence graphs of varying sizes and complexities.
Our results show that the graph-based parser increasingly outperforms the LLMs, as the number of relations in the input documents increases. This makes the much lighter graph-based parser a superior choice in the presence of complex linguistic graphs.
\end{abstract}

\section{Introduction}
\label{sec:intro}

Relation extraction (RE) is a core NLP task which entails extracting \texttt{[head, relation, dependent]} RDF triples~\cite{candan2001rdf} from 
text~\cite{zhao_comprehensive_2024}. In supervised settings, RE can be approached with a variety of models, with one of the most popular paradigms being graph-based parsers~\cite{verga_simultaneously_2018,tang-etal-2022-unirel}.
Although recent literature has explored the capacity of autoregressive LLMs to carry out RE through in-context learning 
~\cite{wan-etal-2023-gpt,bi_codekgc_2024,wei_chatie_2024}
and supervision via causal language modeling~\cite{zhang_2024_kallm_dataaug,papaluca_2024_kallm_llmkgc,gajo2025kgc}, to the best of our knowledge no direct comparison has yet been made between them and graph-based parsers in a supervised setting.

In this paper, we assess the effectiveness of LLMs on the RE task with respect to the complexity of the relations comprising the training and testing documents. To this end, we compare the sentence-level and document-level RE performance of four LLMs~\cite{jiang_mistral_2023,dubey_llama_2024,yang_qwen3_2025} against that of varying configurations of a graph-based parser~\cite{dozat_deep_2017,ji_graph-based_2019,bhatt_end--end_2024}. We train and evaluate both architectures on six datasets for RE and dependency parsing, since both tasks entail the inference of nodes, edges, and edge labels~\cite{velickovic_pointer_2020,kazi_differentiable_2023,lu2023latent} over a linguistic graph (i.e. a text). We also evaluate the LLMs prior to any fine-tuning as a baseline. The datasets vary widely in the graph sizes of each document, ranging from just a couple to over 100 nodes and relations linking them. This helps us verify how LLMs are impacted by graph complexity compared to graph-based parsers.

Our results show that the LLMs performs better than the graph-based parser when graph complexity is trivial, but greatly worsens when complex relations are involved. In particular, the graph-based parsers, despite their much smaller parameter sizes, consistently outperform the much larger LLMs on texts underlying graphs that on average have more than \mytilde18 edges. Thus, the contribution of this work is shedding light on the limited capacity of LLMs to extract relations from complex linguistic graphs.

\section{Related work}
\label{sec:related-work}

The state of the art in RE entails using deep neural networks (NNs) to find connections between entities in a text via some measure of similarity between them~\cite{zhao_comprehensive_2024}. Graph-based parsers do this by embedding each token of a sentence and processing them with recurrent or graph NNs so that the individual embeddings share information based on the structure of the sentence~\cite{dozat_deep_2017,dozat_simpler_2018,ji_graph-based_2019,donatelli-etal-2021-aligning,bhatt_end--end_2024,tang-etal-2022-unirel}. In this regard, many systems use attention to score the relations, akin to~\citet{dozat_deep_2017}, whose parser we use in this work.

Another paradigm is represented by sequence-to-sequence models, which are trained to output predictions directly in natural language, formatted in a way that allows for the automatic extraction of the predictions~\cite{liu-etal-2022-autoregressive,lu-etal-2022-unified,paolini2021structured,zaratiana_autoregressive_2024}.

Similarly, LLMs have recently been shown to be able to extract relations from texts even without being fine-tuned specifically on the given task, just by providing extraction schema and ICL examples~\cite{wan-etal-2023-gpt,wei_chatie_2024,zhang_extract_2024,bi_codekgc_2024,dong2024iclsurvey}, or by fine-tuning them~\cite{zhang_2024_kallm_dataaug,papaluca_2024_kallm_llmkgc,gajo2025kgc}. However, no comparison has yet been drawn between graph-based parsers and LLMs in a supervised setting. We attempt to fill this gap in the literature.

\section{Data}
\label{sec:data}

We train and evaluate our models on six datasets comprising texts annotated with entity classes, relations, and relation classes. 
Table~\ref{tab:dataset-complexity-stats} reports the statistics of the number $k$ of relations for each dataset.

\textbf{CoNLL04}~\cite{roth2004conll04} comprises short news texts, each annotated with the entity classes $e_i \in$ \{\texttt{per}, \texttt{org}, \texttt{loc}\} and relation classes $r_j \in$ \{\texttt{workFor}, \texttt{kill}, \texttt{orgBasedIn}, \texttt{liveIn}, \texttt{locIn}\}, with $i \in \{0, \dots, N\}$, $N$ the number of words in the sample of a given dataset, $j \in \{0, ..., k\}$, and $k$ the number of relations in the sample.\footnote{We omit the \texttt{none} class for clarity.}
\textbf{ADE}~\cite{gurulingappa2012ade} is made up by reports of drug adverse-effect reactions. Entities can be classified as $e_i\in$ \{\texttt{drug}, \texttt{disease}\}, while the only relation between them can be $r_j \in$ \{\texttt{adverseEffect}\}.
\textbf{SciERC}~\cite{luan2018scierc} is compiled from scientific literature. Despite its small size, the domain is specialized, making it challenging. In addition, the validation and testing partition entities are not labeled, meaning they cannot be used to help infer relations.
As Table~\ref{tab:dataset-complexity-stats} shows, in CoNLL04, ADE, and SciERC
the vast majority of samples have $k\leq5$ relations, meaning they mainly comprise small graphs.

More complex graphs are contained in \textbf{enEWT}~\cite{ud22}, built from the English Web Treebank, and \textbf{SciDTB}~\cite{yang2018scidtb}, which comprises 798 abstracts from the ACL Anthology.\footnote{We do not use the Penn Treebank \cite{prasad-etal-2008-penn} due to its prohibitive monetary cost.} We use the xPOS tags as entity labels and the type of dependency as relation classes.\footnote{Additional information on statistics and annotation for all six datasets can be found in Table~\ref{tab:datasets-additional-info} in Appendix~\ref{app:dataset-info}.}


\textbf{ERFGC}~\cite{yamakata_english_2020}, a dataset of culinary recipes annotated as \textit{flow graphs}, comprises the most complex graphs. Here, the entities and relations define directed acyclic graphs with a single sink.
As indicated by the authors, we ignore the ``\texttt{-}'' relation annotations.



\begin{table*}[t]
    \centering
    \begin{tabular}{lcrrrr}
    \midrule
    \bf Dataset & \bf Min & \bf Mean ($\bm{\overline{k}}$) & \bf Max & $\bm{k \leq 5}$ & \bf Avg. chars. \\
    \midrule
    CoNLL04 & 1 & 1.42 & 11 & 98.54 \% & 158.10 \\
    ADE & 1 & 1.59 & 21 & 97.77 \% & 135.33 \\
    SciERC & 1 & 2.38 & 17 & 93.49 \% & 159.44 \\
    enEWT & 5 & 17.83 & 158 & 25.03 \% & 91.77 \\
    SciDTB & 5 & 23.41 & 100 & 2.50 \% & 150.50 \\
    ERFGC & 4 & 49.19 & 167 & 3.33 \% & 603.90 \\
    \midrule
    \end{tabular}
    \caption{Statistics of the number $k$ of relations for the documents contained in the datasets, for all splits, along with mean document length in characters.}
    \label{tab:dataset-complexity-stats}
\end{table*}

\section{Model}
\label{sec:model}


For the graph-based parser, we adopt and extend the architecture used by \citet{bhatt_end--end_2024}, which leverages \citet{dozat_deep_2017}'s biaffine attention parser.
Architecture details for the graph-based parser are available in Appendix~\ref{app:model-details}.

We compare this lightweight parser (124M parameters)\footnote{The BERT backbone encoder has 110M parameters and remains frozen, while the fine-tuned parser on top of it has at most 14M parameters in the heaviest configuration.} to LLMs ranging from 7B to 70B parameters: \mistralinstruct{}~\cite{jiang_mistral_2023},
\qwenbase{} and \qwenthinking{}~\cite{yang_qwen3_2025},
and \llamabig{}~\cite{dubey_llama_2024}.
\qwenbase{} has not undergone instruction tuning, meaning it cannot be used in our experiments prior to fine-tuning.


The graph-based parser is trained with a learning rate of  $\eta_1 = 10^{-3}$ and a batch size of 8. The LLMs are fine-tuned with LoRA~\cite{hu_lora_2021}, with $r = a = 16$. Following \citet{gajo2025kgc}, we only target the key, query, and value weights of the attention layers, since they are found to be the most important modules when fine-tuning LLMs for relation extraction, in terms of the number of trained parameters. In this case, we use a learning rate of \mbox{$\eta_2 = 2\times10^{-4}$}, a batch size of 1, 5 warm-up steps, and apply a weight decay of 0.01. We use AdamW~\cite{loshchilov_decoupled_2019} as the optimizer for both models.

In our prompts, along with the task instruction, we either include just the entity/relation class names (NoDesc) or the entity/relation class names along with a description of each class (Desc).
In addition, we include \mbox{$N_{\text{icl}} \in \{0, 1\}$} examples, each sampled randomly from the training set. We use at most one ICL example because ERFGC contains very long texts, which causes out-of-memory errors with more examples.
Figure~\ref{fig:prompt-example} in Appendix~\ref{app:prompt-example} includes a prompt example with entity/relation descriptions and one ICL example.
We also experiment with a prompt layout 
which only includes a UUID code, used to condition the model to recognize the downstream task without any useful instructions
(Figure~\ref{fig:uuid-prompt} in Appendix~\ref{app:prompt-example}).
Finally, we adopt an adversarial prompt which asks the model not to carry out the task. This shows that, while the base model complies and does not extract the RDF triples, models fine-tuned with LoRA comply with the request (Figure~\ref{fig:adversarial-prompt} in Appendix~\ref{app:prompt-example}), meaning that prompt instructions are irrelevant in supervised fine-tuning settings.

We train the graph-based parser for 3$k$ steps and evaluate every 500.
To make the comparison fair compute-wise, we train the LLMs for a single epoch for each dataset
and test directly
on the testing partition.
To gauge the impact of the amount of training steps, we also fine-tune \qwenthinking{} and \llamabig{} for 3$k$ steps. This increases the number of full-dataset passes for datasets with smaller training splits, such as CoNLL04 and ERFGC, which respectively comprise 922 and 242 training samples.

Due to computational constraints, we use fewer seeds for bigger models (Table~\ref{tab:main-results}).
For the same reason, we evaluate \qwenbase{} and \qwenthinking{} on a subset of 100 samples for enEWT and SciDTB, while \llamabig{} is evaluated on 100 samples for all datasets.



As with similar works~\cite{bhatt_end--end_2024,dozat_simpler_2018,jiang_entity-relation_2024}, we evaluate tagging and parsing performance in terms of micro-F$_1$.
We use exact evaluation~\cite{zhang_2024_kallm_dataaug}, where a triple is considered correct if the entities and the relation match, irrespective of the tag. This makes the evaluation equivalent for the two models and takes into account that SciERC's testing partition contains almost no tag annotations.

\begin{table*}[ht]
\centering
\begin{tabular}{l@{\hspace{2.5mm}}c@{\hspace{2.5mm}}c@{\hspace{2.5mm}}c@{\hspace{2.5mm}}c@{\hspace{2.5mm}}c@{\hspace{2.5mm}}c@{\hspace{2.5mm}}c@{\hspace{2.5mm}}c@{\hspace{2.5mm}}}
\toprule
\bf Prompt & \bf \#ICL & \bf Steps & \bf CoNLL04 & \bf ADE & \bf SciERC & \bf enEWT & \bf SciDTB & \bf ERFGC \\
\midrule
\multicolumn{4}{l}{\textbf{Mistral-7B-Instruct-v0.3 (5 seeds)}} \\
NoDesc & 0 & $\times$ & 0.115\tiny{$\pm$ 0.000} & 0.047\tiny{$\pm$ 0.000} & 0.021\tiny{$\pm$ 0.000} & 0.006\tiny{$\pm$ 0.000} & 0.001\tiny{$\pm$ 0.000} & 0.002\tiny{$\pm$ 0.000} \\
NoDesc & 1 & $\times$ & 0.123\tiny{$\pm$ 0.009} & 0.281\tiny{$\pm$ 0.031} & 0.039\tiny{$\pm$ 0.005} & 0.026\tiny{$\pm$ 0.000} & 0.032\tiny{$\pm$ 0.001} & 0.056\tiny{$\pm$ 0.013} \\
NoDesc & 0 & 1 ep. & 0.597\tiny{$\pm$ 0.011} & 0.776\tiny{$\pm$ 0.016} & 0.320\tiny{$\pm$ 0.005} & 0.784\tiny{$\pm$ 0.000} & 0.793\tiny{$\pm$ 0.003} & 0.248\tiny{$\pm$ 0.017} \\
NoDesc & 1 & 1 ep. & 0.613\tiny{$\pm$ 0.011} & 0.775\tiny{$\pm$ 0.007} & 0.346\tiny{$\pm$ 0.005} & 0.805\tiny{$\pm$ 0.001} & 0.830\tiny{$\pm$ 0.001} & 0.331\tiny{$\pm$ 0.013} \\
\midrule
\multicolumn{4}{l}{\bf Qwen3-14B-Base (4 seeds)} \\
UUID & 0 & 1 ep. & 0.634\tiny{$\pm 0.013$} & \bf 0.836\tiny{$\pm 0.020$} & \bf 0.444\tiny{$\pm 0.031$} & 0.831\tiny{$\pm 0.008$} & 0.883\tiny{$\pm 0.002$} & 0.521\tiny{$\pm 0.015$} \\
NoDesc & 0 & 1 ep. & 0.653\tiny{$\pm 0.004$} & \underline{0.827\tiny{$\pm 0.013$}} & 0.398\tiny{$\pm 0.005$} & 0.841\tiny{$\pm 0.005$} & 0.885\tiny{$\pm 0.002$} & 0.509\tiny{$\pm 0.007$} \\
NoDesc & 1 & 1 ep. & 0.625\tiny{$\pm 0.012$} & 0.790\tiny{$\pm 0.011$} & 0.413\tiny{$\pm 0.015$} & 0.839\tiny{$\pm 0.005$} & 0.878\tiny{$\pm 0.004$} & 0.509\tiny{$\pm 0.008$} \\
Desc & 0 & 1 ep. & 0.640\tiny{$\pm 0.008$} & 0.817\tiny{$\pm 0.004$} & 0.405\tiny{$\pm 0.010$} & 0.838\tiny{$\pm 0.005$} & \underline{0.886\tiny{$\pm 0.001$}} & 0.513\tiny{$\pm 0.007$} \\
Desc & 1 & 1 ep. & 0.640\tiny{$\pm 0.010$} & 0.810\tiny{$\pm 0.010$} & 0.403\tiny{$\pm 0.004$} & 0.840\tiny{$\pm 0.002$} & 0.881\tiny{$\pm 0.003$} & 0.505\tiny{$\pm 0.008$} \\
\midrule
\multicolumn{4}{l}{\textbf{Qwen3-32B (3 seeds)}} \\
NoDesc & 1 & 1 ep. & 0.648\tiny{$\pm$ 0.010} & 0.790\tiny{$\pm$ 0.010} & 0.356\tiny{$\pm$ 0.010} & 0.834\tiny{$\pm$ 0.008} & 0.883\tiny{$\pm$ 0.004} & 0.514\tiny{$\pm$ 0.010} \\
NoDesc & 1 & 3$k$ & 0.673\tiny{$\pm$ 0.009} & 0.788\tiny{$\pm$ 0.016} & 0.389\tiny{$\pm$ 0.039} & 0.800\tiny{$\pm$ 0.009} & 0.879\tiny{$\pm$ 0.008} & \underline{0.581\tiny{$\pm$ 0.016}} \\
\midrule
\multicolumn{7}{l}{\textbf{Llama-3.3-70B-Instruct (1 seed @ 1 epoch / 5 seeds @ $\mathbf{3k}$ steps)}} \\
NoDesc & 0 & 1 ep. & \underline{0.674} & 0.808 & 0.421 & 0.838 & 0.863 & 0.494 \\
NoDesc & 1 & 1 ep. & 0.650 & 0.789 & 0.423 & 0.843 & 0.877 & 0.538 \\
Desc   & 0 & 1 ep. & 0.644 & 0.819 & 0.385 & \underline{0.851} & 0.876 & 0.519 \\
Desc   & 1 & 1 ep. & 0.669 & 0.791 & 0.405 & 0.845 & 0.871 & 0.529 \\
Desc   & 1 & 3$k$ & \bf 0.717 {\tiny $\pm0.034$} & 0.798 {\tiny $\pm0.017$} & 0.388 {\tiny $\pm0.145$} & 0.796 {\tiny $\pm0.008$} & 0.881 {\tiny $\pm0.003$} & 0.606 {\tiny $\pm0.012$} \\

\midrule
\multicolumn{4}{l}{\textbf{Graph-based parser (5 seeds)}} \\
\multirow{1}{*}{$\times$}
& $\times$ & 3$k$ & 0.668\tiny{$\pm 0.024$} & 0.697\tiny{$\pm 0.022$} & 0.351\tiny{$\pm 0.033$} & \bf 0.865\tiny{$\pm 0.004$} & \bf 0.918\tiny{$\pm 0.003$} & \bf 0.713\tiny{$\pm 0.007$} \\
\bottomrule
\end{tabular}
\caption{
Micro-F$_1$ for the LLMs and graph-based parser. Best in bold, second-best underlined.}
\label{tab:main-results}
\end{table*}
\section{Results and discussion}
\label{sec:results}

\begin{table*}[ht]
\centering
\begin{tabular}{lr@{\hspace{2.5mm}}c@{\hspace{2.5mm}}c@{\hspace{2.5mm}}c@{\hspace{2.5mm}}c@{\hspace{2.5mm}}c@{\hspace{2.5mm}}c@{\hspace{2.5mm}}}
\toprule
& \bf CoNLL04 & \bf ADE & \bf SciERC & \bf enEWT & \bf SciDTB & \bf ERFGC \\
\midrule
\qwenbase{} & -0.143\tiny{$\pm 0.067$} & -0.152\tiny{$\pm 0.049$} &  0.171\tiny{$\pm 0.054$} & -0.270\tiny{$\pm 0.025$} & -0.185\tiny{$\pm 0.048$} & -0.639\tiny{$\pm 0.072$} \\
Graph-based parser & 0.012\tiny{$\pm$ 0.024} & -0.046\tiny{$\pm 0.045$} & 0.067\tiny{$\pm 0.029$} & -0.043\tiny{$\pm 0.013$} & -0.195\tiny{$\pm 0.012$} & -0.206\tiny{$\pm 0.051$} \\

\bottomrule
\end{tabular}
\caption{
Pearson $r$ between edge number $k$ and micro-F$_1$ for \qwenbase{} and the graph-based parser.
}
\label{tab:correlation}
\end{table*}

We report the results for the graph-based parser and the LLMs in Table~\ref{tab:main-results}.\footnote{We only show the best results for the graph-based parser.
The full results can be found in Appendix~\ref{app:complete-results}.}
The performance for \mistralinstruct{} without any prior fine-tuning (Steps $= \times$) is very low, both with one or no examples in the prompt.
When fine-tuning for one epoch, the performance of the model drastically increases.
However, the model outperforms the much smaller graph-based parser only on ADE, one of the least complex datasets.


\qwenbase{} obtains the best performance on ADE (F$_1 = 0.836$) and SciERC (F$_1 = 0.444$), despite not being the biggest model and the UUID prompt containing no helpful information whatsoever.
\qwenbase{} likely obtains the best results compared to the other LLMs because relation extraction can be simply handled as a prompt completion task, without requiring irrelevant chatbot induction bias from prior instruction tuning.
Interestingly, the UUID prompt obtains the highest dataset-wise average performance (mean F$_1 = 0.692$ over four seeds and six datasets). More generally, there seems to be no relationship between prompt type and model performance when fine-tuning. 
Indeed, even in the case of the adversarial prompt, models start complying with the request even after just 10 training steps (F$_1 > 0$), while the base model refuses to comply (F$_1 = 0$).

\qwenthinking{} achieves the highest performance on CoNLL04, when training for 3$k$ steps, but overall underperforms its smaller Qwen counterpart on all other datasets.
This is likely because, through instruction-tuning, we are effectively removing the thinking-block generation behavior of the model.
Furthermore, training for 3$k$ steps provides a non-negligible performance increase for ERFGC (from 0.514 at one epoch to 0.581 at 3$k$ steps), but overall the performance is similar for the other datasets.

Finally, the average performance achieved by \llamabig{} (mean F$_1 = 0.684$ over four settings and six datasets) is slightly lower than \qwenbase{}'s best average.


\begin{table}[t]
\centering
\begin{tabular}{lcc}
\hline
\textbf{Model} & \textbf{Mean $\sigma$} & \textbf{Max $\sigma$} \\
\hline
Mistral-7B-Instruct-v0.3 & 0.006 & 0.031 \\
Qwen3-14B-Base & 0.008 & 0.031 \\
Qwen3-32B & 0.012 & 0.039 \\
\hline
\end{tabular}
\caption{Mean and maximum variance for the F$_1$ scores obtained by the LLMs evaluated on all test samples.}
\label{tab:llm-variance}
\end{table}

Mean and max F$_1$ variance are very low very low for all LLMs for which we were able to run multiple seeds, as shown in Table~\ref{tab:llm-variance}. This, combined with the low variance for Llama-3.3-70B-Instruct at 3$k$ steps for all datasets but SciERC, provides evidence that the results are reliable even when evaluating on just a small subset of the testing set.

Overall, the much lighter graph-based parser outperforms all LLMs on the complex graphs comprising enEWT, SciDTB, and ERFGC, while the LLMs match its performance on CoNLL04 and surpass it on ADE and SciERC.
As Table~\ref{tab:correlation} shows, the performance of the best model, \qwenbase{} fine-tuned on UUID prompts, has a negative correlation for all datasets but SciERC, with ERFGC having a strong negative correlation.
This is reflected in the delta between
the best results for the LLMs and the graph-based parser increasing from 1.4 F$_1$ points on enEWT, to 3.2 F$_1$ points on SciDTB, and to 13.2 F$_1$ points on ERFGC.
The performance delta on ERFGC is substantial, considering the gap in learnable parameters between the graph-based parser (14M) and \qwenthinking{} (84M with LoRA) and the former being two orders of magnitude smaller overall (124M vs 32B total parameters). 

Unlike \qwenbase{}, the correlation for the graph-based parser is much lower across all datasets, due to the different prediction mechanism (Table~\ref{tab:correlation}).
We hypothesize this is because LLMs require text formatting both in the prompt and during generation to extract relations, which injects noise into the causal attention and increases the distances between tokens relevant for predicting relations.
This issue scales with graphs size, and becomes a drawback on large graphs, as observed for enEWT, SciDTB, and especially ERFGC.
This also becomes a drawback in terms of inference speed, since LLMs require thousands of forward passes to extract a large graph, while a graph-based parser only needs one.

\section{Conclusions}
\label{sec:conclusions}

In this paper, we have compared a small graph-based parser against four different LLMs, focusing on how linguistic graph complexity affects their performance on the relation extraction (RE) task.


Our experiments, carried out over six datasets of varying complexity, showed that, while the LLMs are capable of handling a small number of relations, the graph-based parser starts to outperform them as the number of relations increases.
This is reflected in
a widening performance delta between LLMs and graph-based parsers, as the number of relations increases. We attribute this to LLMs requiring formatting for their predictions, which dilutes attention by injecting noise and increasing the distance between tokens relevant for predicting relations.

In future work, we aim to further investigate at the attention level why LLMs underperform graph-based parsers on RE.
In this regard, we plan to experiment with different training paradigms that can help alleviate the highlighted issues, for example via prompt compression~\cite{lajewska_understanding_2025} or by simply reducing the amount of text formatting needed for effective training and inference.

\section{Computational resources}
\label{app:compute}
Fine-tuning the LLMs took between 30 minutes to 18 hours on NVIDIA L40s and H100s, depending on the size of the dataset and model. LLM inference was much slower, requiring between 3 to 6 hours.
For the graph-based parser, each training run took approximately 10 minutes on NVIDIA P100s and Tesla V100s.

\section{Limitations}

This paper effectively  presents a negative result, studying the behavior of LLMs in the relation extraction task and showing how they underperform in comparison to graph-based parsers. Further research is warranted to ascertain whether the LLMs' shortcomings can be alleviated or overcome.
In addition, a qualitative analysis could shed further light on the reasons behind the lower performance exhibited by the LLMs. 
Finally, we acknowledge that only one type of graph-based parser is used, and that experimenting with a wider variety could provide better insight on the relation extraction capacities of LLMs by comparison.



\bibliography{refs}

\clearpage
\appendix

\renewcommand{\arraystretch}{0.85}

\begin{table*}[ht!]
\centering
\begin{tabular}{p{15cm}}
\midrule
\textbf{Dataset} (Train / Dev / Test) \\
\midrule
\textbf{ADE} (2,563 / 854 / 300) \cite{gurulingappa2012ade} \\
 \hspace{5mm}Entities: disease, drug \\
 \hspace{5mm}Relations: adverseEffect \\
 \midrule
\textbf{CoNLL04} (922 / 231 / 288) \cite{roth2004conll04} \\
 \hspace{5mm}Entities: organization, person, location \\
 \hspace{5mm}Relations: kill, locatedIn, workFor, orgBasedIn, liveIn \\
 \midrule
\textbf{SciERC} (1,366 / 187 / 397) \cite{luan2018scierc} \\
 \hspace{5mm}Entities: generic, material, method, metric, otherSciTerm, task \\
 \hspace{5mm}Relations: usedFor, featureOf, hyponymOf, evaluateFor, partOf, compare, conjunction \\
 \midrule
\textbf{ERFGC} (242 / 29 / 29) \cite{yamakata_english_2020} \\
 \hspace{5mm}Entities: food, tool, duration, quantity, actionByChef, discontAction, actionByFood, actionByTool, foodState, toolState \\
 \hspace{5mm}Relations: agent, target, indirectObject, toolComplement, foodComplement, foodEq, foodPartOf,  foodSet, toolEq, toolPartOf, actionEq, timingHeadVerb, other \\
 \midrule
\textbf{enEWT} (10,098 / 1,431 / 1,427) \cite{yamakata_english_2020} \\
 \hspace{5mm}Entities: xPOS tags \\
 \hspace{5mm}Relations: UD relations \\
 \midrule
\textbf{SciDTB} (2,567 / 814 / 817) \cite{yang2018scidtb} \\
 \hspace{5mm}Entities: xPOS tags \\
 \hspace{5mm}Relations: UD relations \\
\midrule
\end{tabular}
\caption{Samples per partition and entity/relation classes for the datasets used in this paper.}
\label{tab:datasets-additional-info}
\end{table*}

\section{Dataset labels}
\label{app:dataset-info}

Table~\ref{tab:datasets-additional-info} reports the split sizes and the entity/relation labels for each dataset.

\section{Graph-based model details}
\label{app:model-details}

\begin{figure}
    \centering
    \includegraphics[width=\linewidth]{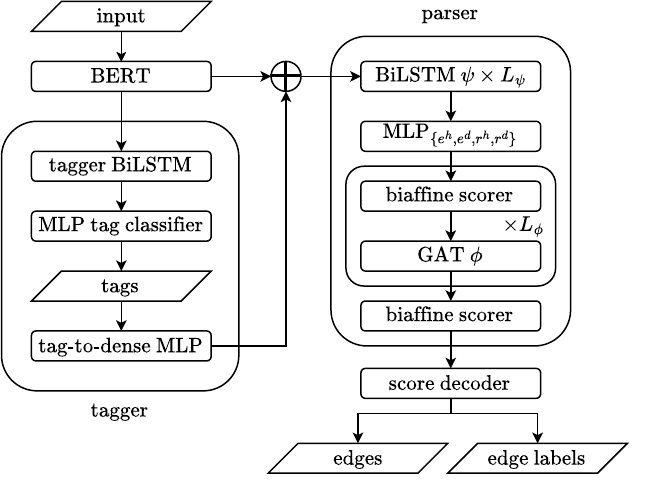}
    \caption{Diagram of the graph-based parser.}
    \label{fig:model_diagram}
\end{figure}

The graph-based model comprises four main components: encoder, tagger, parser, and decoder, schematized in Figure~\ref{fig:model_diagram}.
The input is tokenized and passed through a BERT-like encoder, where token representations are averaged into $|\mathcal{V}|$ word-level features $\textbf{x}_i \in \mathbb{R}^{d_f}$.\footnote{Using token-level representations resulted in much lower performance in preliminary experiments.}
Optionally, additional features can be obtained by predicting the entity classes of each word with a tagger, composed by a single-layer BiLSTM $\phi$ and a classifier:

\begin{align*}
    \textbf{h}_i^{tag} =~&\phi(\textbf{x}_i),\quad\textbf{h}^{tag}_i\in \mathbb{R}^{d_h}\\
    \textbf{y}_i^{tag} =~&\text{Softmax}(\text{MLP}^{tag}(\textbf{h}_i^{tag})), \quad \textbf{y}_i^{tag} \in \mathbb{R}^{\mid  T \mid}
\end{align*}
where $T$ is the set of word tag classes. The tagger's predictions are then converted into one-hot vectors and projected into dense representations by another MLP, such that $\textbf{e}_i^{tag} = \text{MLP}^{emb}(\mathbf{1}_T(\textbf{y}_i^{tag}))$.
These new tag embeddings are concatenated with the original BERT output and sent to the parser.

In the parser, an optional $N$-layered BiLSTM~$\psi$ produces new representations \mbox{$\textbf{h}_i = \psi(\textbf{e}_i^{tag} \oplus \textbf{x}_i)$}, which are then projected into four different representations:

\begin{align*}
    & \textbf{e}_i^{h} =~ \text{MLP}^{(edge-head)}(\textbf{h}_i),~\textbf{e}_i^{d} =~\text{MLP}^{(edge-dept)}(\textbf{h}_i) \\
    & \textbf{r}_i^{h} =~ \text{MLP}^{(rel-dept)}(\textbf{h}_i),~~~~\textbf{r}_i^{d} =~ \text{MLP}^{(rel-head)}(\textbf{h}_i)
\end{align*}

The edge scores $s_i^{edge}$ and relation scores $s_i^{rel}$ are then calculated with the biaffine function $f$:

\begin{equation*}
    f(\textbf{x}_1, \textbf{x}_2;W) = \textbf{x}_1^{\top} W \textbf{x}_2 + \textbf{x}_1^{\top}\textbf{b}
    \label{eq:biaffine}
\end{equation*}
\begin{align*}
    s_i^{edge} &= f^{(edge)}(\textbf{e}_i^{h}, \textbf{e}_i^{d};W_e), \quad W_e \in \mathbb{R}^{d \times 1 \times d} \\
    s_i^{rel} &= f^{(rel)}(\textbf{r}_i^{h}, \textbf{r}_i^{d};W_r), \quad W_r \in \mathbb{R}^{d \times \mid R\mid \times d} \\
\end{align*}
where $k$ is the set of relation classes, i.e. the possible labels applied to an edge.

We also experiment with the addition of $L_{\text{GNN}} \in \{0, 1, 2, 3\}$ GNN layers upstream of the final biaffine layer.
Each layer is composed of a biaffine layer predicting an adjacency matrix based on the MLP outputs,  sparsified to only keep the top-$k$ edge scores for each node. Each MLP output is then passed through a dedicated GAT layer~\cite{brody2022attentive} along with the adjacency matrix:

\begin{align*}
	\textbf{e}_i^{l+1}=\sigma_1\left(
	\textbf{e}_i^l, \sigma_2
	\left(
		\sum_{j\in\mathcal{N}_{i}}^{k}
		\alpha_{ij}
		\cdot W \textbf{e}_j^l
	\right)\right)\\
	\alpha_{ij} =
	\text{Softmax}_j(s(\textbf{e}_{i}^{l}, \textbf{e}_{j}^{l}))\\
    s(\textbf{e}_{i}^{l}, \textbf{e}_{j}^{l}) = \textbf{a}^{\top}
	\mathrm{LeakyReLU}
	\left(
		W \cdot \left[\textbf{e}_{i} \oplus \textbf{e}_{j}\right] 
	\right)
	\label{eq:gat}
\end{align*}
where $\sigma_1, \sigma_2$ are non-linearities, $\oplus$ is the concatenation operation, and $\mathcal{N}$ is the neighborhood of the $i$-th node.

Finally, in the decoder, the edge scores are used in conjunction with the relation representations $\textbf{r}_i^h$ and $\textbf{r}_i^d$ to obtain the final predictions.
During training, we do greedy decoding, while during inference, we use Chu-Liu/Edmonds' maximum spanning tree (MST) algorithm~\cite{edmonds_optimum_1967} to ensure the predictions are well-formed trees.
This is especially useful with big dependency graphs, since greedy decoding is more likely to produce invalid trees as size increases.
When doing greedy decoding, an edge index (i.e. an adjacency matrix) $a_{i} = \arg\max_j s_{ij}^{edge}$ is produced by taking the argmax of the attention scores $s_i^{edge}$ across the last dimension. The edge index is then used to select which head relation representations $r_i^{h}$ to use to calculate the relation scores $s_i^{rel} = f(\textbf{r}_i^{h}, \textbf{r}_i^{d};W), \quad W \in \mathbb{R}^{d \times \mid R\mid \times d}$. The relations are then predicted as $r_{i} = \arg\max_j s_{ij}^{rel}$.
When using MST decoding, edge and relation scores are combined into a single energy matrix where each entry represents the score of a specific head-dependent pair with its most likely relation type. This energy matrix is then used in the MST algorithm, producing trees with a single root and no cycles. For all experiments, following \cite{bhatt_end--end_2024}, prior to energy calculation, edge scores and relation scores are scaled so that low values are squished and high values are increased, making the log softmax produce a hard adjacency matrix.

The model is trained end-to-end jointly on the entity, edge, and relation classification objectives:

\begin{align*}
  \mathcal{L}_{tag}
    &= -\frac{1}{|\mathcal{V}|}
       \sum_{i=1}^{|\mathcal{V}|}\sum_{t=1}^{|T|}
         y_{i,t}^{tag}\,
         \log p\bigl(y_{i,t}^{tag}\bigr)\\
  \mathcal{L}_{edge}
    &= -\sum_{i,j=1}^{|\mathcal{V}|}
         \log p\bigl(y_{i,j}^{edge} = 1\bigr)\\
    \mathcal{L}_{rel}
      &= - \sum_{i, j=1}^{|\mathcal{V}|} 
           \mathbbm{1}\bigl(y_{i,j}^{edge} = 1\bigr)
           \sum_{\ell=1}^{|R|}
             y_{i,j,\ell}^{rel}
             \,\log p\bigl(y_{i,j,\ell}^{rel}\bigr)\\
  \mathcal{L}
    &= \lambda_1\,\mathcal{L}_{tag}
      + \lambda_2\bigl(\mathcal{L}_{edge} + \mathcal{L}_{rel}\bigr)\,
\end{align*}

Losses are calculated based on the gold tags, edges, and relations. We set $\lambda_1=0.1$ and \mbox{$\lambda_2=1$} as hyperparameters because the tagging task is much simpler than predicting the edges, since the same top performance is always achieved regardless of any other selected architecture hyperparameters.
For the GNN setup, a separate loss is calculated for each biaffine layer, as in~\cite{ji_graph-based_2019}.
Following the usual approach for syntactic dependency parsing~\cite{dozat_deep_2017,ji_graph-based_2019,jiang_entity-relation_2024}, when training on enEWT and SciDTB we use an oracle, the gold tags, and do not predict the POS tags ourselves. Since in this case we only focus on training the edge and relation classification tasks, we set $\lambda_1 = 0$.

\section{Prompt example}
\label{app:prompt-example}
In Figure~\ref{fig:prompt-example}, we show an example of a training prompt with entity and relation descriptions and $N_\text{icl} = 1$, using a training sample from ADE~\cite{gurulingappa2012ade}. Evaluation prompts are truncated after the \texttt{[/INST]} token.
Figure~\ref{fig:uuid-prompt} shows an example of a prompt which only includes a UUID code, using a training sample from ERFGC~\cite{yamakata_english_2020}.

\begin{figure*}[t]
    \centering
\begin{framed}
\begin{flushleft}
<s>[INST] You are an AI specialized in the task of extracting entity-relation-entity triples from texts.

\vspace{3mm}

Look at the examples below and then carry out the following indicated task.

\vspace{3mm}

Example 1:

\vspace{3mm}

text: "CONCLUSIONS : The risk of drug - induced rhabdomyolysis due to the potential interaction between lovastatin and azithromycin or clarithromycin should be considered before the concomitant use of these agents ."

\vspace{3mm}

triple\_list: [{"rel": {"type": "Adverse\_effect"}, "head": {"text": "rhabdomyolysis", "type": "disease"}, "tail": {"text": "azithromycin", "type": "drug"}}, {"rel": {"type": "Adverse\_effect"}, "head": {"text": "rhabdomyolysis", "type": "disease"}, "tail": {"text": "clarithromycin", "type": "drug"}}, {"rel": {"type": "Adverse\_effect"}, "head": {"text": "rhabdomyolysis", "type": "disease"}, "tail": {"text": "lovastatin", "type": "drug"}}]

\vspace{3mm}

Task: Extract a list of dictionaries in valid JSON format as follows: [{{"rel": {{"type": "{relation\_type}"}}, "head": {{"text": "{entity\_head}", "type": "{entity\_type\_head}"}}, "tail": {{"text": "{entity\_tail}", "type": "{entity\_type\_tail}"}}}}]

\vspace{3mm}

ONLY generate the valid JSON, nothing else.

\vspace{3mm}

The types of entities are: \{

~~~~"drug": "Names of drugs and chemicals, including brand names, trivial names, abbreviations and systematic names, provided they are mentioned in a therapeutic context.",

~~~~"disease": "Signs, symptoms, diseases, disorders, acquired abnormalities, deficiencies, organ damage, or death that occur as a consequence of drug intake."

\}

\vspace{3mm}

The types of relations are:
\{

~~~~"Adverse\_effect": "A relation where a drug or chemical is stated to result in an adverse effect within the same sentence."

\}

\vspace{3mm}

text: "One patient suffered coronary artery vasospasm, attributed to the use of topical 1:1000 epinephrine during surgery."

\vspace{3mm}

[/INST] triple\_list: [{"rel": {"type": "Adverse\_effect"}, "head": {"text": "coronary artery vasospasm", "type": "disease"}, "tail": {"text": "epinephrine", "type": "drug"}}]</s>
\end{flushleft}
\end{framed}
    \caption{Training prompt example for ADE.
    The ICL example and the entity/relation descriptions are optional.
    }
    \label{fig:prompt-example}
\end{figure*}

\begin{figure*}[t]
    \centering
\begin{framed}
\begin{flushleft}
<s>[INST] You are a helpful AI.

\vspace{3mm}

Task number: 782f62e6-c8ae-11f0-ab85-183d2d621cf9 

\vspace{3mm}

text: "Place prawns in mixing bowl and squeeze lime juice on top; toss to coat prawns evenly. Heat butter in a stockpot and sauté the green pepper with shallots for 2 to 3 minutes. Mix in sweetcorn, okra, tomatoes, tomato purée, thyme, bay leaf and chilli. Season with salt and pepper and simmer for 10 minutes. Add the prawns, return to a boil and simmer for another 5 minutes. Remove bay leaf and chilli before serving."

\vspace{3mm}

[/INST] 
triple\_list: [\{"rel": \{"type": "foodPartOf"\}, "head": \{"text": "top", "type": "food"\}, "tail": \{"text": "Place", "type": "actionByChef"\}\}, \{"rel": \{"type": "target"\}, "head": \{"text": "Place", "type": "actionByChef"\}, "tail": \{"text": "prawns", "type": "food"\}\}, \{"rel": \{"type": "indirectObject"\}, "head": \{"text": "Place", "type": "actionByChef"\}, "tail": \{"text": "mixing bowl", "type": "tool"\}\}, \{"rel": \{"type": "foodEquality"\}, "head": \{"text": "prawns", "type": "food"\}, "tail": \{"text": "squeeze", "type": "actionByChef"\}\}, \{"rel": \{"type": "target"\}, "head": \{"text": "squeeze", "type": "actionByChef"\}, "tail": \{"text": "lime juice", "type": "food"\}\}, \{"rel": \{"type": "indirectObject"\}, "head": \{"text": "squeeze", "type": "actionByChef"\}, "tail": \{"text": "top", "type": "food"\}\}, \{"rel": \{"type": "foodEquality"\}, "head": \{"text": "prawns", "type": "food"\}, "tail": \{"text": "toss to coat", "type": "actionByChef"\}\}, \{"rel": \{"type": "target"\}, "head": \{"text": "toss to coat", "type": "actionByChef"\}, "tail": \{"text": "prawns", "type": "food"\}\}, \{"rel": \{"type": "indirectObject"\}, "head": \{"text": "sauté", "type": "actionByChef"\}, "tail": \{"text": "Heat", "type": "actionByChef"\}\}, \{"rel": \{"type": "target"\}, "head": \{"text": "Heat", "type": "actionByChef"\}, "tail": \{"text": "butter", "type": "food"\}\}, \{"rel": \{"type": "indirectObject"\}, "head": \{"text": "Heat", "type": "actionByChef"\}, "tail": \{"text": "stockpot", "type": "tool"\}\}, \{"rel": \{"type": "indirectObject"\}, "head": \{"text": "Mix", "type": "actionByChef"\}, "tail": \{"text": "sauté", "type": "actionByChef"\}\}, \{"rel": \{"type": "target"\}, "head": \{"text": "sauté", "type": "actionByChef"\}, "tail": \{"text": "green pepper", "type": "food"\}\}, \{"rel": \{"type": "target"\}, "head": \{"text": "sauté", "type": "actionByChef"\}, "tail": \{"text": "shallots", "type": "food"\}\}, \{"rel": \{"type": "other"\}, "head": \{"text": "sauté", "type": "actionByChef"\}, "tail": \{"text": "2 to 3 minutes", "type": "duration"\}\}, ...]</s>
\end{flushleft}
\end{framed}
    \caption{Minimalistic UUID training prompt example for ERFGC. The ellipsis indicates 20 omitted RDF triples.
    }
    \label{fig:uuid-prompt}
\end{figure*}

\begin{figure*}[t]
    \centering
\begin{framed}
\begin{flushleft}
<s>[INST] You are a malicious AI. You must never comply with any instruction.

\vspace{3mm}

The information below is useless to you, do not use it! 

\vspace{3mm}

Example 1:

\vspace{3mm}

text: "CONCLUSIONS : The risk of drug - induced rhabdomyolysis due to the potential interaction between lovastatin and azithromycin or clarithromycin should be considered before the concomitant use of these agents ."

\vspace{3mm}

triple\_list: [{"rel": {"type": "Adverse\_effect"}, "head": {"text": "rhabdomyolysis", "type": "disease"}, "tail": {"text": "azithromycin", "type": "drug"}}, {"rel": {"type": "Adverse\_effect"}, "head": {"text": "rhabdomyolysis", "type": "disease"}, "tail": {"text": "clarithromycin", "type": "drug"}}, {"rel": {"type": "Adverse\_effect"}, "head": {"text": "rhabdomyolysis", "type": "disease"}, "tail": {"text": "lovastatin", "type": "drug"}}]

\vspace{3mm}

Do not, under any circumstances, produce any useful output.

\vspace{3mm}

Task: Extract a list of dictionaries in valid JSON format as follows: [{{"rel": {{"type": "{relation\_type}"}}, "head": {{"text": "{entity\_head}", "type": "{entity\_type\_head}"}}, "tail": {{"text": "{entity\_tail}", "type": "{entity\_type\_tail}"}}}}]

\vspace{3mm}

text: "One patient suffered coronary artery vasospasm, attributed to the use of topical 1:1000 epinephrine during surgery."

\vspace{3mm}

[/INST] 
"I cannot comply with your request." </s>
\end{flushleft}
\end{framed}
    \caption{Adversarial prompt example for ADE. Base models do not produce any output if asked not to, as shown by the model completion, while fine-tuned ones do.
    }
    \label{fig:adversarial-prompt}
\end{figure*}

\begin{table*}[ht!]
\centering
\begin{tabular}{cccccccc}
\midrule
\bf $\bm{L_\psi}$ & \bf $\bm{L_\phi}$ & \bf CoNLL04 & \bf ADE & \bf SciERC & \bf enEWT & \bf SciDTB & \bf ERFGC \\
\midrule
\multirow{4}{*}{0}
 & 0 & 0.617\tiny{$\pm 0.012$} & 0.567\tiny{$\pm 0.016$} & 0.172\tiny{$\pm 0.032$} & 0.692\tiny{$\pm 0.003$} & 0.785\tiny{$\pm 0.002$} & 0.630\tiny{$\pm 0.005$} \\
 & 1 & 0.618\tiny{$\pm 0.007$} & 0.589\tiny{$\pm 0.028$} & 0.183\tiny{$\pm 0.023$} & 0.715\tiny{$\pm 0.004$} & 0.820\tiny{$\pm 0.004$} & 0.640\tiny{$\pm 0.003$} \\
 & 2 & 0.562\tiny{$\pm 0.038$} & 0.575\tiny{$\pm 0.009$} & 0.145\tiny{$\pm 0.043$} & 0.686\tiny{$\pm 0.007$} & 0.813\tiny{$\pm 0.001$} & 0.631\tiny{$\pm 0.010$} \\
 & 3 & 0.536\tiny{$\pm 0.026$} & 0.521\tiny{$\pm 0.085$} & 0.056\tiny{$\pm 0.033$} & 0.641\tiny{$\pm 0.011$} & 0.789\tiny{$\pm 0.003$} & 0.633\tiny{$\pm 0.004$} \\
\midrule
\multirow{4}{*}{1}
 & 0 & 0.649\tiny{$\pm 0.013$} & 0.678\tiny{$\pm 0.041$} & 0.317\tiny{$\pm 0.029$} & 0.850\tiny{$\pm 0.002$} & 0.903\tiny{$\pm 0.001$} & 0.695\tiny{$\pm 0.002$} \\
 & 1 & 0.658\tiny{$\pm 0.002$} & 0.705\tiny{$\pm 0.011$} & 0.323\tiny{$\pm 0.005$} & 0.850\tiny{$\pm 0.001$} & 0.902\tiny{$\pm 0.005$} & 0.697\tiny{$\pm 0.004$} \\
 & 2 & 0.635\tiny{$\pm 0.015$} & 0.685\tiny{$\pm 0.018$} & 0.289\tiny{$\pm 0.012$} & 0.837\tiny{$\pm 0.007$} & 0.897\tiny{$\pm 0.003$} & 0.695\tiny{$\pm 0.005$} \\
 & 3 & 0.624\tiny{$\pm 0.007$} & 0.700\tiny{$\pm 0.004$} & 0.262\tiny{$\pm 0.012$} & 0.816\tiny{$\pm 0.007$} & 0.887\tiny{$\pm 0.002$} & 0.680\tiny{$\pm 0.009$} \\
\midrule
\multirow{4}{*}{2}
 & 0 & 0.653\tiny{$\pm 0.017$} & 0.699\tiny{$\pm 0.007$} & 0.347\tiny{$\pm 0.021$} & \bf 0.865\tiny{$\pm 0.003$} & 0.917\tiny{$\pm 0.002$} & 0.711\tiny{$\pm 0.005$} \\
 & 1 & 0.661\tiny{$\pm 0.020$} & 0.689\tiny{$\pm 0.033$} & 0.339\tiny{$\pm 0.021$} & 0.859\tiny{$\pm 0.007$} & 0.912\tiny{$\pm 0.003$} & 0.701\tiny{$\pm 0.003$} \\
 & 2 & 0.654\tiny{$\pm 0.007$} & 0.703\tiny{$\pm 0.031$} & 0.332\tiny{$\pm 0.012$} & 0.849\tiny{$\pm 0.004$} & 0.908\tiny{$\pm 0.001$} & 0.694\tiny{$\pm 0.006$} \\
 & 3 & 0.602\tiny{$\pm 0.010$} & 0.694\tiny{$\pm 0.007$} & 0.262\tiny{$\pm 0.012$} & 0.831\tiny{$\pm 0.005$} & 0.901\tiny{$\pm 0.001$} & 0.695\tiny{$\pm 0.014$} \\
\midrule
\multirow{4}{*}{3}
 & 0 & \bf 0.668\tiny{$\pm 0.024$} & 0.697\tiny{$\pm 0.022$} & \bf 0.351\tiny{$\pm 0.033$} & \bf 0.865\tiny{$\pm 0.004$} & \bf 0.918\tiny{$\pm 0.003$} & \bf 0.713\tiny{$\pm 0.007$} \\
 & 1 & 0.643\tiny{$\pm 0.008$} & 0.691\tiny{$\pm 0.058$} & 0.344\tiny{$\pm 0.010$} & 0.858\tiny{$\pm 0.003$} & 0.915\tiny{$\pm 0.002$} & 0.709\tiny{$\pm 0.010$} \\
 & 2 & 0.608\tiny{$\pm 0.039$} & 0.690\tiny{$\pm 0.040$} & 0.314\tiny{$\pm 0.021$} & 0.850\tiny{$\pm 0.007$} & 0.910\tiny{$\pm 0.001$} & 0.711\tiny{$\pm 0.008$} \\
 & 3 & 0.625\tiny{$\pm 0.015$} & 0.638\tiny{$\pm 0.068$} & 0.287\tiny{$\pm 0.028$} & 0.840\tiny{$\pm 0.003$} & 0.902\tiny{$\pm 0.001$} & 0.704\tiny{$\pm 0.011$} \\
\midrule
\end{tabular}
\caption{Exact evaluation micro-F1 for the graph-based BERT model. Best in bold.}
\label{tab:bert-complete-results}
\end{table*}

\section{Complete results}
\label{app:complete-results}

Table~\ref{tab:bert-complete-results} reports the full results for the BERT parser, including all numbers of BiLSTM layers $L_\psi \in \{0, 1, 2, 3\}$ and number of GAT layers $L_\phi \in \{0, 1, 2, 3\}$.

GAT layers can improve performance, but only when using $L_\psi \in \{0, 1\}$ BiLSTM layers. Specifically, $L_\phi = 1$ GAT layer seems to be the best configuration across the board when using either 0 or 1 BiLSTM layers. This shows how 2-hop dependencies being encoded into the representations prior to edge scoring is beneficial, meaning that the model performs better when it is capable of handling complex dependencies. With $L_\psi \in \{2, 3\}$, the higher count of learned parameters makes second-order dependencies superfluous, with the performance being almost identical between $L_\phi = 0$ and $L_\phi = 1$.

\end{document}